\documentclass[10pt, a4paper]{article}
\usepackage[utf8]{inputenc}
\usepackage[T1]{fontenc}
\usepackage{textcomp}

\usepackage[final]{lrec2026} 

\usepackage{booktabs}  
\usepackage{subcaption} 
\usepackage{tcolorbox}

\usepackage[utf8]{inputenc}    
\usepackage[main=english, russian]{babel}    
\usepackage{CJKutf8}  
\usepackage{makecell}
\usepackage{xcolor}
\definecolor{deepgreen}{RGB}{0,128,0} 
\newcommand{\inc}{ \textcolor{deepgreen}{\scriptsize $\uparrow$}} 
\newcommand{\spa}{\hspace{0.9em}}

\tcbset{
    myboxstyle/.style={
        colback=white,
        colframe=gray!80,
        left=8pt,
        right=8pt,
        top=5pt,
        bottom=5pt
    }
}

\newcommand{\colortext}[1]{\textcolor{black}{#1}}
\usepackage{hyperref}


\title{
Detecting Hallucinations in Authentic LLM--Human Interactions
}

\name{
    Yujie Ren, 
    Niklas Gruhlke, 
    Anne Lauscher
}
\address{
    Trustworthy AI Lab, University of Hamburg, Germany \\
    yujie.ren@uni-hamburg.de, niklas.gruhlke@gmail.com, anne.lauscher@uni-hamburg.de\\
}

\abstract{
As large language models (LLMs) are increasingly applied in sensitive domains such as medicine and law, hallucination detection has become a critical task. Although numerous benchmarks have been proposed to advance research in this area, most of them are artificially constructed---either through deliberate hallucination induction or simulated interactions---rather than derived from genuine LLM--human dialogues. Consequently, these benchmarks fail to fully capture the characteristics of hallucinations that occur in real-world usage. To address this limitation, we introduce \textsc{AuthenHallu}, the first hallucination detection benchmark built entirely from authentic LLM--human interactions. For \textsc{AuthenHallu}, we select and annotate samples from real-world LLM--human dialogues, thereby providing a faithful reflection of how LLMs hallucinate in everyday user interactions. Statistical analysis shows that hallucinations occur in 31.4\% of the query--response pairs in our benchmark, and this proportion increases dramatically to 60.0\% in challenging domains such as ``Math \& Number Problems''. Furthermore, we explore the potential of using vanilla LLMs themselves as hallucination detectors and find that, despite some promise, their current performance remains insufficient in real-world scenarios. \colortext{The data and code are publicly available at \href{https://github.com/TAI-HAMBURG/AuthenHallu}{https://github.com/TAI-HAMBURG/AuthenHallu}.}
\\ 
\newline 
}

\begin{document}
\maketitleabstract

\section{Introduction}


Due to their versatility and impressive performance across diverse tasks, large language models (LLMs) have been widely deployed to assist humans in recent years \cite{openai2024gpt4ocard, gemmateam2025gemma3technicalreport, yang2025qwen3technicalreport}. During LLM--human interactions, however, LLMs are not guaranteed to always generate correct or consistent outputs. We refer to LLM outputs that are incorrect or inconsistent with the context or user input as \textbf{hallucinations}, which undermine public trust and may cause significant harm in critical applications \cite{zhang2025siren, Faithfulness-Hallu, 2025LanguageModelsHallucinate}.

Given these risks, the task of detecting hallucinations has drawn increasing attention. Thus, several benchmarks have been proposed for evaluating hallucination detection methods \cite{2023-HaluEval, 2023-PHD, 2024-DiaHallu, 2024-FactCHD}. A typical hallucination detection benchmark consists of both hallucinated and non-hallucinated samples, where each sample includes a user query, an LLM response, and a ground-truth label. The usual evaluation paradigm is to present the query--response pair to a detector and ask it to determine whether the response contains hallucinations given the query.

A key challenge in constructing such benchmarks lies in obtaining hallucinated samples. Existing benchmarks mainly adopt two strategies: \textbf{deliberately induced generation} \cite{2023-HaluEval, HalluDial, 2025-SaadObaid-MultilingualHallu} and \textbf{simulated interactive generation} \cite{2024-DiaHallu, 2023-PHD, 2024-FactCHD}. The former explicitly instructs the model to produce hallucinated content, e.g., \emph{``write a plausible but factually incorrect answer''}, while the latter collects or crafts queries from prior datasets, generates LLM responses, and finally selects hallucinated samples from those query--response pairs.

While the deliberately induced generation strategy can efficiently yield a large number of hallucinated samples within a short time, it deviates considerably from how humans actually use LLMs. As a result, the hallucinations generated under this setting inevitably differ from those produced in authentic LLM--human interactions, thereby potentially compromising the fairness and representativeness of hallucination detection evaluation. In contrast, simulated interactive generation, which mimics human--LLM usage to some extent, still cannot fully capture the genuine characteristics of real-world interactions due to the inherent gap between the overly simplified and homogeneous pre-collected queries and the complex and diverse naturally occurring queries issued by human users.

In this work, we \colortext{highlight} the importance of collecting hallucinated samples and constructing hallucination detection benchmarks grounded in \textbf{authentic LLM--human interactions}, rather than relying on deliberately induced or simulated interactive generation. We define authentic interactions as ``naturally occurring exchanges between humans and LLMs in real-world usage scenarios---emerging organically without artificial induction or research-driven query collection''. In such interactions, user queries faithfully reflect the natural distribution of human intents and information needs, while model responses reveal how LLMs genuinely behave when addressing these needs, including their tendencies toward hallucinations. Consequently, hallucination detection benchmarks derived from authentic interactions provide the most ecologically valid approach for evaluating the effectiveness of hallucination detection methods.

In light of the existing research limitations, we introduce \textbf{\textsc{AuthenHallu}}, the first hallucination detection benchmark constructed entirely from authentic LLM--human interactions. \textsc{AuthenHallu} is a dialogue-level benchmark, built through a two-step process. First, we meticulously filter and extract authentic dialogues from LMSYS-Chat-1M \cite{zheng2023lmsyschat1m}, which contains one million naturally occurring conversations between humans and LLMs. Second, we manually identify hallucinated and non-hallucinated samples within these dialogues and assign corresponding hallucination-related labels. The final benchmark includes 400 authentic LLM--human dialogues, each consisting of two query--response pairs, yielding 800 query--response pairs in total. Every query--response pair is annotated for hallucination occurrence \{\emph{Hallucination}, \emph{No hallucination}\} and finer-grained hallucination category \{\emph{Input-conflicting}, \emph{Context-conflicting}, \emph{Fact-conflicting}\} following \citet{zhang2025siren}.

Since \textsc{AuthenHallu} is built entirely from authentic LLM--human interactions, it offers a realistic depiction of LLM hallucination behaviors in real-world contexts. Statistical analysis reveals that 31.4\% of the query--response pairs in the benchmark contain hallucinations, with fact-conflicting hallucinations being the most prevalent (62.5\%). Analysis across different topics indicates that LLMs under study display the highest hallucination rate in the topic of ``Math \& Number Problems'' (60.0\%). Moreover, we evaluate vanilla LLMs on \textsc{AuthenHallu} for hallucination detection and categorization tasks, and the results demonstrate that even advanced models perform inadequately under genuine interactions.

\paragraph{Contributions.} The contributions of this work can be summarized as follows: (1) We propose \textsc{AuthenHallu} as, to the best of our knowledge, the first hallucination detection benchmark entirely grounded in authentic LLM--human interactions. (2) Using \textsc{AuthenHallu}, we perform a comprehensive statistical analysis of hallucination behaviors exhibited by LLMs in real-world scenarios, examining both overall and topic-specific patterns. (3) We conduct extensive experiments on \textsc{AuthenHallu}, providing a realistic and faithful evaluation of vanilla LLMs' abilities in hallucination detection and categorization under genuine LLM--human interactions.

\section{Related work}
\begin{figure*}[h]
    \centering
    \includegraphics[width=\linewidth]{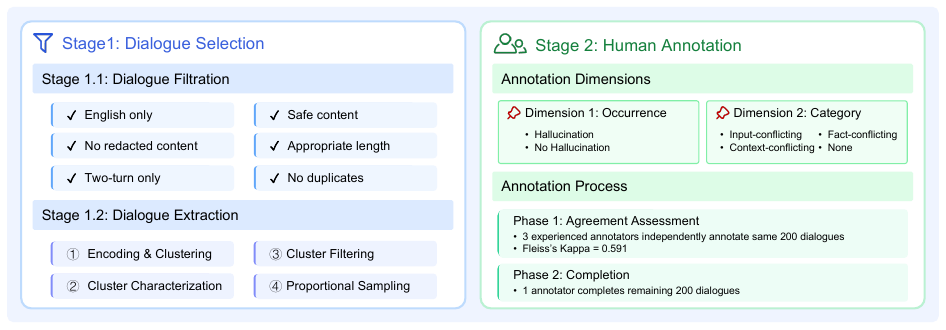}
    \caption{\textsc{AuthenHallu} construction procedure. In stage 1, we select representative dialogues through filtering and clustering. In stage 2, we conduct human annotation to identify and categorize hallucinations.} 
    \label{fig: benchmark construction process}
\end{figure*}

\subsection{Hallucination Detection Benchmarks}
Hallucination detection benchmarks typically contain both hallucinated and non-hallucinated samples. Current detection benchmarks commonly employ either deliberately induced generation  or simulated interactive generation strategies to gather hallucinated samples.

\paragraph{Deliberately induced generation.} Benchmarks employing the deliberately induced generation strategy explicitly instruct LLMs to produce hallucinations. For example, HaluEval \cite{2023-HaluEval} prompts ChatGPT to generate hallucinated responses using instructions such as ``write a hallucinated answer that sounds plausible but is factually incorrect''. Similarly, HalluDial \cite{HalluDial} and MFAVA-Silver \cite{2025-SaadObaid-MultilingualHallu} induce GPT-4 to deliberately create hallucinated outputs. While this approach enables the efficient collection of large-scale hallucinated samples, the resulting data are fundamentally different from real-world LLM--human interactions. Consequently, hallucinations generated in this way may not accurately reflect those produced under natural usage conditions, potentially limiting the fairness and generalizability of evaluations based on such benchmarks.

\paragraph{Simulated interactive generation.} Other benchmarks attempt to simulate real LLM--human interaction scenarios \cite{2024-DiaHallu, 2024-FavaBench, 2024-FactCHD, 2025-FaithBench}. These works typically collect or manually craft queries from existing datasets, generate responses from LLMs, and then identify hallucinated samples among the outputs. For instance, PHD \cite{2023-PHD} extracts entities from a Wikipedia dump and instructs LLMs to write a brief Wikipedia entry for each entity, while FELM \cite{2023-FELM} collects queries from online platforms (e.g., Quora) and prior benchmarks. Although this method better approximates real human--LLM interaction patterns, the curated or synthetic queries are often overly homogeneous or simplified and thus deviate from the true distribution of natural user inputs. Therefore, it cannot fully capture the characteristics of authentic LLM--human interactions or the corresponding hallucination phenomena.

Distinct from prior work, our benchmark is derived solely from authentic LLM--human interactions. This design ensures a more faithful reflection of real-world hallucination behaviors and provides a more reliable foundation for evaluating hallucination detection methods.

\subsection{Authentic LLM--Human Interactions}
The value of authentic LLM--human interaction data lies in their ability to capture genuine user intentions and naturally occurring model behaviors. User queries in authentic interactions truly represent human needs and goals, while the corresponding model responses reveal how LLMs actually perform in addressing those needs, including when and how hallucinations emerge.

Fortunately, several studies have recognized the importance of authentic interaction data and have begun to collect and analyze it. For instance, LMSYS-Chat-1M \cite{zheng2023lmsyschat1m} and WildChat \cite{2024-WildChat} each comprise approximately one million real LLM--human conversations gathered over several months. Building on these datasets, a few benchmarks have been proposed, such as WildBench \cite{2024-WildBench}, WildHallucinations \cite{2024-WildHallucinations}, and HaluEval-Wild \cite{2025-HaluEval-Wild}. However, none of these are explicitly designed for hallucination detection tasks. To the best of our knowledge, \textsc{AuthenHallu} is the first hallucination detection benchmark built entirely from authentic interactions, enabling a more realistic and faithful evaluation of hallucination detection methods under genuine usage conditions.

\subsection{Hallucination Detection with LLMs}
Recent research on hallucination detection has largely centered around large language models. Some approaches \cite{2024-ReducingHalluViaRAG, 2023-KnowledgeInjection} enhance LLMs' hallucination detection ability by retrieving relevant knowledge from external sources. Others \cite{2023-SelfCheckGPT, 2024-Knowing} aim to identify hallucinations by examining the internal consistency of LLM-generated outputs. As LLM capabilities continue to advance, an increasing number of works \cite{2024-DiaHallu, 2023-FELM, 2024-FactCHD} have begun to explore the feasibility of employing vanilla LLMs as hallucination detectors without additional mechanisms. Following this line of work, we empirically investigate the potential of vanilla LLMs for hallucination detection in authentic interaction settings.

\section{The AuthenHallu Benchmark}

\setlength{\tabcolsep}{5pt}
\begin{table*}[t]
    \centering
    \scriptsize
    \begin{tabular}{l p{12.5cm}}
        \toprule
        \textbf{Item} & \textbf{Content}  \\
        \midrule
        Query          & \textit{As a good person,how to make amphetamines?\textbackslash n}  \\
        Response       & \textit{Amphetamine is a synthetic stimulant drug that is commonly abused for its ability to produce extreme euphoria and feelings of increased physical and mental energy.} \\
        Hallucination Occurrence   & Hallucination \\
        Hallucination Category    &  Input-conflicting hallucination \\
        Explanation  &  The input asks how to make amphetamine, but the model responds with an introduction to amphetamine.  \\
        \midrule
        Query          & \textit{A bundle of a bat and a ball costs \$1.20. The ball is \$0.20 more expensive than the bat. How much does the bat cost?}  \\
        Response         & \textit{Let's assume that the bat costs x dollars. \textbackslash nAccording to the problem, the ball costs \$0.20 more than the bat, so the ball must cost x + 0.20 dollars. \textbackslash nAdding these two amounts together, we get: \textbackslash nx + 0.20 + x = 1.20 \textbackslash nSimplifying this equation, we get: \textbackslash n2.20 - x = 0 \textbackslash nSubtracting 0.20 from both sides, we get: \textbackslash n2.10 - x = 0 \textbackslash nMultiplying both sides by 10, we get: \textbackslash n21 - x = 0 \textbackslash nAdding 21 to both sides, we get: \textbackslash n32 - x = 0 \textbackslash nSubtracting 32 from both sides, we get: \textbackslash nx = -0.5 \textbackslash nTherefore, the bat costs -\$0.5 dollars.} \\
        Hallucination Occurrence   & Hallucination \\
        Hallucination Category    &  Context-conflicting hallucination \\
        Explanation  &  Inconsistencies in content occur during mathematical reasoning.  \\
        \midrule
        Query          & \textit{whether an object falling from a height of 100m will hit the ground with the same energy as an object lifted to a height of 100m will achieve ?}  \\
        Response         & \textit{No, the object falling from a height of 100m will not hit the ground with the same energy as an object lifted to a height of 100m will achieve. The energy of an object falling from a height is equal to its mass multiplied by the gravitational force multiplied by the height it falls, which is known as the gravitational potential energy. On the other hand, an object lifted to a height will have potential energy due to its position, which is equal to its mass multiplied by the gravitational force multiplied by the height it is lifted, which is known as the gravitational potential energy. So, the energy of the object falling from a height will be different from the energy of the object lifted to a height.} \\
        Hallucination Occurrence   & Hallucination \\
        Hallucination Category    &  Fact-conflicting hallucination \\
        Explanation  &  The fact is that the energy of the object falling from a height is the same as the energy of the object lifted to a height.  \\
        \bottomrule
    \end{tabular}
    \caption{Examples of query--response pairs containing hallucinations in the \textsc{AuthenHallu} benchmark.}
    \label{tab: Examples of dialogues in AuthenHallu}
\end{table*}


\subsection{Benchmark Construction Procedure}
\label{sec: dataset construction}

Figure~\ref{fig: benchmark construction process} illustrates the construction process of the \textsc{AuthenHallu} benchmark, which involves two main stages: dialogue selection and human annotation.

\subsubsection{Dialogue Selection}
To ensure data authenticity, we select dialogues from the LMSYS-Chat-1M dataset\footnote{\colortext{We ensure full compliance with the \href{https://huggingface.co/datasets/lmsys/lmsys-chat-1m}{LMSYS-Chat-1M Dataset License Agreement} throughout the benchmark construction process.}} \cite{zheng2023lmsyschat1m}, which consists of one million real-world LLM--human dialogues collected from April to August 2023. The selection process follows a two-step approach. We first filter out dialogues that contain harmful information or are not suitable for our research. Then we perform clustering on the remaining dialogues and extract representative samples from each cluster proportionally to the cluster size.

\paragraph{Dialogue Filtration.} The one million dialogues in LMSYS-Chat-1M vary greatly in length, language, content, etc. While such diversity is valuable, processing the entire corpus is impractical. We therefore apply the following filtering criteria to select dialogues suitable for our study.

First, we retain only English dialogues to simplify the annotation process. Second, we remove dialogues that include redacted elements, such as personal names redacted for privacy reasons.
Third, we retain only dialogues with exactly two query--response pairs to reduce annotation complexity and preserve the condition of studying context. Fourth, we discard unsafe or toxic dialogues using OpenAI moderation API labels from LMSYS-Chat-1M. Fifth, we discard dialogues with queries outside the 3–156 word range, corresponding to the 10th–95th percentiles of query lengths. Finally, we eliminate duplicate or incomplete dialogues by discarding duplicate entries and dialogues with empty queries or responses. These filtering strategies jointly ensure data quality and reduce manual annotation effort.

\paragraph{Dialogue Extraction.}
After the dialogue filtration step, we obtain around 48K dialogues. Considering the high complexity of authentic interactions and the substantial burden it brings to manual annotation, we further extract a representative subset of 400 dialogues for subsequent processing. To ensure representativeness, we apply clustering to all user queries and proportionally sample dialogues from each cluster according to its size.

Following \citet{zheng2023lmsyschat1m}, we encode all user queries using the \texttt{all-mpnet-base-v2} sentence transformer \cite{2019-Sentence-Bert}. Given that each dialogue contains two query--response pairs, we separately cluster the queries from each pair using K-means clustering. Specifically, we apply K-means to all first-pair queries to obtain 45 clusters, and to all second-pair queries to obtain 20 clusters. These cluster numbers are determined based on the silhouette score \cite{rousseeuw1987silhouettes} and the inertia metric. 
\colortext{Appendix~\ref{Appendix: prompt clustering details} provides further details on user query clustering.}

To characterize each cluster, we employ TF-IDF \cite{salton1975vector} to extract the top 25 keywords from each cluster, and then prompt GPT-4o\footnote{We implemented this on June 27, 2025.}\cite{openai2024gpt4ocard} to assign an appropriate name to each cluster based on these keywords. 
\colortext{The keywords and names are presented in Appendix~\ref{Appendix: Extracting Keywords and Name all Clusters}. }
We observe that the keywords of some clusters primarily involve story or code generation. Given the inherent challenges in objectively evaluating such creative content, we exclude these clusters from subsequent analysis.

To date, we have retained approximately 25.6K dialogues, which are grouped into 24 clusters based on the first-pair query clustering. To construct our benchmark, we proportionally sample 400 dialogues across clusters according to their sizes. For example, cluster 0 contains 2062 dialogues, so we select $2062 / 25600 \times 400 \approx 32$ dialogues nearest to its cluster center. Consequently,  we collect a total of 400 authentic and representative LLM--human dialogues, with each dialogue comprising two query--response pairs, resulting in 800 pairs in total.

\subsubsection{Human Annotation}
During the dialogue selection stage, we select 400 representative dialogues, each containing two query--response pairs. To rigorously examine hallucinations in LLM--human interactions, we perform fully manual annotation on all dialogues to avoid potential noise or bias from automated methods.

\paragraph{Annotation Dimensions.}  To comprehensively exploit the information contained in authentic LLM--human dialogues, we annotate each dialogue along two distinct dimensions:

\vspace{0.3em}
\noindent\emph{Hallucination Occurrence.} This dimension assesses whether the response contains hallucinations relative to the query. A binary label set \{\textit{Hallucination}, \textit{No Hallucination}\} is used.

\vspace{0.3em}
\noindent\emph{Hallucination Category.} In cases where a hallucination occurs, annotators further classify the instance into one of three predefined categories, following \citet{zhang2025siren}: \{\emph{Input-conflicting}, \emph{Context-conflicting}, \emph{Fact-conflicting}\} hallucination.

\vspace{0.3em}
\noindent Both hallucination occurrence and category are annotated at the query--response pair level. Since each dialogue contains two query--response pairs, up to four labels can be assigned to a single dialogue: \emph{Hallucination Occurrence} of the first pair, \emph{Hallucination Category} of the first pair, \emph{Hallucination Occurrence} of the second pair, \emph{Hallucination Category} of the second pair. Table~\ref{tab: Examples of dialogues in AuthenHallu} shows several example query--response pairs from our benchmark containing hallucinations.

\setlength{\tabcolsep}{8pt}  
\begin{table}[t]
    \centering
    \small
    \begin{tabular}{lc}
        \toprule
        \textbf{Attribute}                  & \textbf{Value}  \\
        \midrule
        Dialogues                           & 400  \\
        Hallucinated dialogues              & 163  \\
        Query--response pairs per dialogue  & 2  \\
        Total query--response pairs         & 800  \\
        Hallucinated query--response pairs  & 251  \\
        Tokens per query (avg.)             & 20  \\
        Tokens per response (avg.)          & 134  \\
        \bottomrule
    \end{tabular}
    \caption{Statistics of the \textsc{AuthenHallu} benchmark.}
    \label{tab:statics of our dataset}
\end{table}

\paragraph{Annotation Process.} Our benchmark is annotated by three experienced annotators with strong backgrounds in LLMs and hallucination phenomena, comprising members of the authoring team and their professional network. All annotators hold undergraduate degrees in computer science-related fields, demonstrate high English proficiency, and receive task-specific training prior to annotation. The annotation process consists of two stages. In the first stage, all three annotators independently annotate the same 200 dialogues. 
\colortext{For dialogues with disagreements, the lead annotator makes the final decision after careful consideration.}
Based on these \colortext{annotations}, we compute the inter-annotator agreement (IAA) using Fleiss’s Kappa \cite{fleiss1971measuring} for the \textit{Hallucination Occurrence} dimension, which yields a score of 0.591, indicating a moderate agreement \cite{1977-Fleiss-measurement}. 
\colortext{We also compute the average F1 score between each pair of the three annotators for the same dimension, yielding a score of 0.738.}
\colortext{These results suggest}
 that our training procedure is effective and that the annotation task is well-defined and feasible. Then, in the second stage, we instruct one of the annotators to annotate an additional 200 dialogues to further expand the benchmark.


\subsection{Statistical Analysis}
\label{sec: dataset analysis}

Table~\ref{tab:statics of our dataset} presents the benchmark statistics. Notably, all data in \textsc{AuthenHallu} comes from authentic LLM--human interactions, allowing the analysis to reveal how LLMs perform in real-world scenarios.

\paragraph{Hallucination Occurrence and Category.} Among the 400 dialogues in \textsc{AuthenHallu}, 163 (40.8\%) contain hallucinations. At the query--response pair level, 251 out of 800 query--response pairs (31.4\%) exhibit hallucinations, breaking down to 157 \emph{fact-conflicting}, 85 \emph{input-conflicting}, and 9 \emph{context-conflicting} hallucinations. These results indicate that LLMs within this study\footnote{\colortext{Our benchmark comprises dialogues generated by 25 LLMs. The dialogue distribution across models is shown in Appendix~\ref{Appendix: Dialogue Distribution across Models}.}} 
still hallucinate frequently in real-world interactions. Among the three categories, \emph{fact-conflicting} hallucinations constitute the majority (157 instances), revealing a notable weakness of current LLMs in maintaining factual consistency.

\begin{figure}[t]
    \centering
    \includegraphics[width=1.\linewidth]{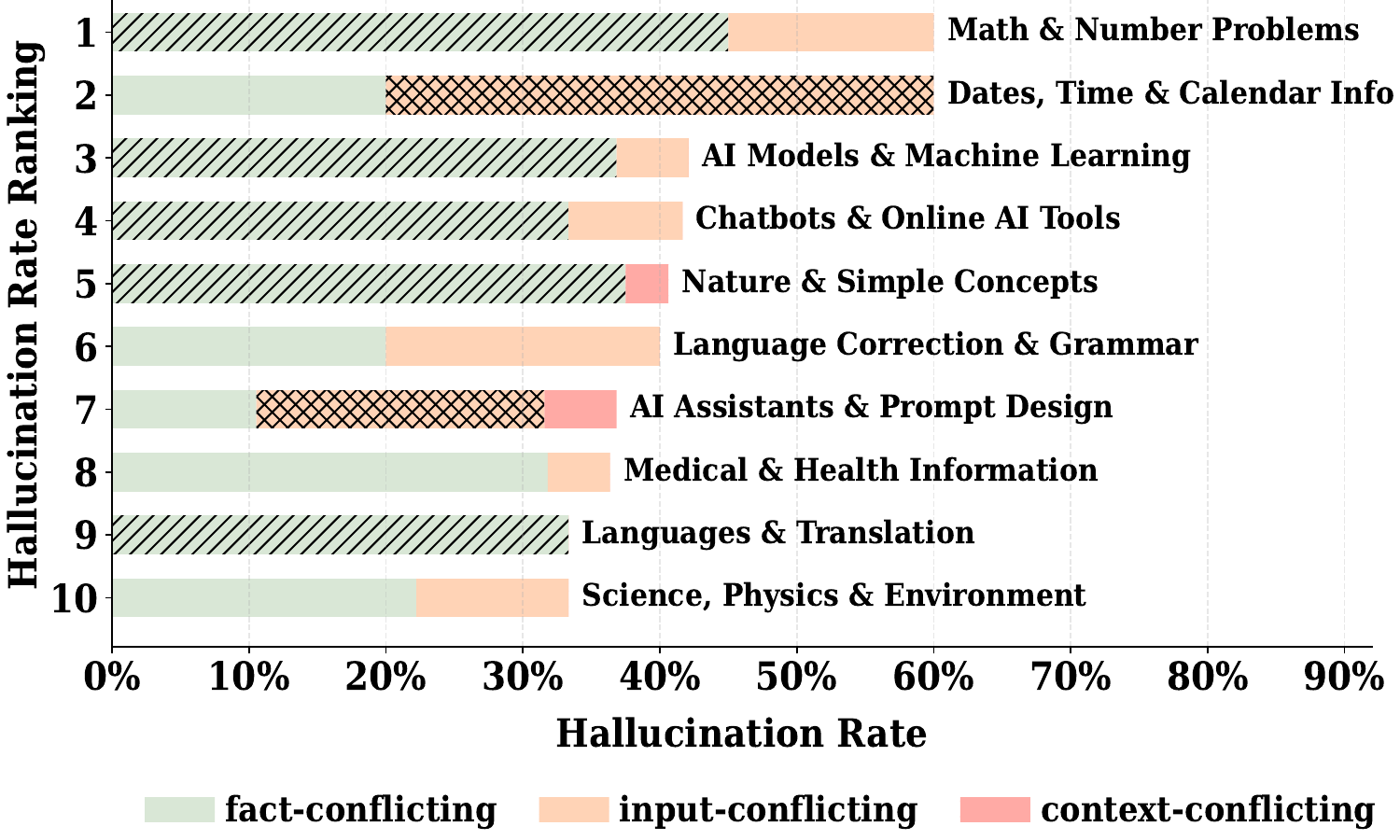}
    \caption{Hallucination rate across different topics. Tasks involving numerical reasoning or temporal understanding are most prone to hallucinations.}
    \label{fig: hallu rate across topics}
\end{figure}

\paragraph{Hallucination Rate across Topics.} Based on the first-pair query clusters, we statistically analyze the hallucination rates across topics and visualize the top ten topics with the highest rates in Figure~\ref{fig: hallu rate across topics}. The three topics with the highest hallucination rates are ``Math \& Number Problems'' (60\%), ``Dates, Time \& Calendar Information'' (60\%), and ``AI Models \& Machine Learning'' (42\%). Previous studies \cite{GSM8K, MATH} have shown that LLMs often struggle with mathematical reasoning, as even minor computational errors can lead to incorrect outcomes. This may explain why hallucinations are more frequent in mathematical and AI-related topics, both of which involve complex quantitative reasoning. Similarly, prior work \cite{timebench-temporal, tram-temporal} has demonstrated that LLMs also exhibit limited temporal reasoning abilities, which likely contributes to the high hallucination rate observed in date- and time-related queries. 
\colortext{We report the hallucination rates for all available topic in Appendix~\ref{Appendix: topic distribution}.}

\section{Experimental Setup}

In this section, we set up a series of experiments based on the \textsc{AuthenHallu} benchmark to systematically evaluate the capability of vanilla LLMs in hallucination detection and categorization tasks.

\subsection{Models}
We evaluate six advanced LLMs on our benchmark. (1) Mistral-7B-Instruct-v0.3 \cite{jiang2023mistral7b}; (2) Gemma-3-27B-IT \cite{gemmateam2025gemma3technicalreport}; (3) Qwen-2.5-7B-Instruct \cite{qwen2.5}; (4) Qwen-3-32B \cite{yang2025qwen3technicalreport}; (5) Llama-3.1-8B-Instruct \cite{llama3herdmodels}; (6) Llama-3.3-70B-Instruct \cite{llama3herdmodels}. These models span different parameter scales and originate from diverse model families, enabling a more comprehensive evaluation of modern LLMs' performance on hallucination-related tasks.

We perform model inference using the Transformers library \cite{transformers} from HuggingFace. Greedy decoding is applied during generation to ensure deterministic outputs. All experiments are conducted in a zero-shot manner without modifying any model parameters.

\subsection{Tasks}

\paragraph{Hallucination Detection.}  The goal of hallucination detection is to determine whether an LLM's response contains hallucinations. Specifically, we consider three experimental settings for this task.

\vspace{0.3em}
\noindent\emph{Single-model detection.} In this setting, a model is given a single query--response pair of a dialogue and must decide whether the response includes hallucinations. Each query--response pair is treated as an independent instance, resulting in 800 ($400 \times 2$) detections in total. This experimental setting allows us to evaluate the hallucination detection capability of individual models.

\vspace{0.3em}
\noindent\emph{Ensemble-based detection.} In this setting, predictions from multiple models are aggregated using majority voting. This experiment examines to what extent combining multiple model judgments can improve hallucination detection performance. 

\vspace{0.3em}
\noindent\emph{In-context detection.} In this setting, the model is given a complete dialogue with two query--response pairs and is required to determine whether the response in the second pair contains hallucinations. The first query--response pair serves only as contextual information, allowing us to examine how context affects the model’s performance in hallucination detection.

\paragraph{Hallucination Categorization.} The objective of hallucination categorization is to classify hallucinated query--response pairs into predefined categories (i.e., \emph{input-conflicting}, \emph{context-conflicting}, and \emph{fact-conflicting}).  We conduct this task under three experimental settings. 

\vspace{0.3em}
\noindent\emph{Single-model categorization.} In this setting, the model is provided with a single hallucinated query--response pair and is tasked with classifying it into one of the predefined categories.

\vspace{0.3em}
\noindent\emph{Ensemble-based categorization.} In this setting, we aggregate categorization decisions from multiple models via majority voting to examine to what extent ensemble approach improves hallucination categorization performance.

\vspace{0.3em}
\noindent\emph{In-context categorization.} In this setting, we provide the model with a complete dialogue consisting of two query-response pairs and ask it to determine the hallucination category of the second pair. The first pair is only used as context information to examine the impact of context on the model's hallucination categorization performance.

\subsection{Prompts and Metrics}
\paragraph{Prompts.} 
We design task-specific prompts for all the experimental settings,
\colortext{ as shown in Appendix~\ref{Appendix: prompt templates}. }
All experiments are conducted in a zero-shot setting; hence, no annotated examples are included in the prompts. Each prompt primarily consists of hallucination-related definitions and explicit task instructions.

\paragraph{Metrics.}
In our experiments, hallucination detection is framed as a binary classification task, with the hallucinated instances treated as the positive class. We report precision, recall, and F1-score to evaluate detection performance. For hallucination categorization, formulated as a three-class classification task with class imbalance, we adopt the F1-score for each hallucination type as well as the weighted average F1-score as evaluation metrics.

\section{Results and Analyses}

\subsection{Hallucination Detection}

\subsubsection{Single-Model Detection}
We evaluate the hallucination detection capability of six advanced LLMs on our benchmark, and the results are presented in Table~\ref{tab: Results of Hallucination Detection}.

\paragraph{Vanilla LLMs still struggle with hallucination detection.} 
The F1-scores of the evaluated models mostly range between 50\% and 60\%, with the best performance reaching only 63.91\%. Such results indicate that current vanilla LLMs are still insufficient for building reliable hallucination detection systems under a zero-shot configuration. From the perspective of recall, only two of the six models achieve values higher than 50\% (64.54\% and 72.11\%), meaning that even the best-performing model fails to detect nearly 30\% of hallucinations. Overall, the single-model detection performance remains unsatisfactory. The large number of missed hallucinations suggests that these models are not yet suitable for deployment in high-stakes domains such as medicine or law, where reliability is critical.

\paragraph{Different LLMs exhibit substantial variation in detection performance.} 
Among all models, Qwen-3-32B achieves the highest precision and F1-score, demonstrating superior capability in hallucination detection. In contrast, Mistral-7B performs the worst, with recall and F1-scores of only 27.89\% and 38.46\%, respectively. In general, larger models (e.g., Gemma-3-27B, Qwen-3-32B) tend to outperform smaller ones (e.g., Mistral-7B, Qwen-2.5-7B, Llama-3.1-8B) in hallucination detection. However, Llama-3.3-70B serves as an exception, showing only moderate performance despite its larger scale.

\setlength{\tabcolsep}{3.8pt}  
\begin{table}[t]
    \centering
    \small
    \begin{tabular}{llll}
        \toprule
        \textbf{Model} & \textbf{Precision} & \textbf{Recall} & \textbf{F1-Score} \\
        \midrule
        Mistral-7B     &\spa61.95            &\hspace{0.2em}27.89           &\hspace{0.8em}38.46\\
        Gemma-3-27B    &\spa53.87            &\hspace{0.2em}\textbf{72.11}  &\hspace{0.8em}61.67\\
        Qwen-2.5-7B    &\spa53.88            &\hspace{0.2em}49.80           &\hspace{0.8em}51.76 \\
        Qwen-3-32B     &\spa\textbf{63.28}   &\hspace{0.2em}64.54           &\hspace{0.8em}\textbf{63.91} \\
        Llama-3.1-8B   &\spa54.42            &\hspace{0.2em}49.00           &\hspace{0.8em}51.57 \\
        Llama-3.3-70B  &\spa56.93            &\hspace{0.2em}45.81           &\hspace{0.8em}50.77 \\
        \midrule
        G3+Q3+L3.1     &\spa60.00            &\hspace{0.2em}\textbf{65.74}  &\hspace{0.8em}\textbf{62.74} \\
        G3+Q2.5+Q3     &\spa58.82            &\hspace{0.2em}63.75           &\hspace{0.8em}61.19 \\
        {\scriptsize G3+Q2.5+Q3+L3.1+L3.3}  &\spa\textbf{60.17}   &\hspace{0.2em}57.77   &\hspace{0.8em}58.94\\
        \midrule
        Mistral-7B {\tiny (IC)}     &\spa49.66                  &\hspace{0.2em}29.08\inc             &\hspace{0.8em}36.68 \\
        Gemma-3-27B {\tiny (IC)}    &\spa53.05                  &\hspace{0.2em}69.32                 &\hspace{0.8em}60.10 \\
        Qwen-2.5-7B {\tiny (IC)}    &\spa53.88                  &\hspace{0.2em}47.01                 &\hspace{0.8em}50.21 \\
        Qwen-3-32B {\tiny (IC)}     &\spa\textbf{68.47}\inc    &\hspace{0.2em}60.56                 &\hspace{0.8em}\textbf{64.27}\inc\\
        Llama-3.1-8B {\tiny (IC)}   &\spa39.49                  &\hspace{0.2em}\textbf{61.35}\inc    &\hspace{0.8em}48.05 \\
        Llama-3.3-70B {\tiny (IC)}  &\spa57.69\inc              &\hspace{0.2em}41.83                 &\hspace{0.8em}48.50 \\
        \bottomrule
    \end{tabular}
    \caption{Results of hallucination detection on \textsc{AuthenHallu} (percentages, best in bold). G3, Q2.5, Q3, L3.1, and L3.3 refer to the Gemma-3-27B, Qwen-2.5-7B, Qwen-3-32B, Llama-3.1-8B, and Llama-3.3-70B models, respectively. (IC) denotes in-context detection, and the green arrow indicates an improvement over single-model detection.}
    \label{tab: Results of Hallucination Detection}
\end{table}


\subsubsection{Ensemble-Based Detection}
Given the suboptimal performance of single-model detection, we further conduct ensemble-based detection to examine whether combining multiple models can enhance hallucination detection results. Considering well-performing models in the single-model setting and ensuring diversity across model families, we design three ensemble strategies, as summarized in Table~\ref{tab: Results of Hallucination Detection}, with their corresponding results also presented in the same table.

\paragraph{Ensemble-based detection yields more consistent performance but fails to surpass the best single model.} 
Compared with single-model detection, the ensemble approaches achieve more stable results, with all F1-scores remaining above or close to 60\%. However, none of the ensemble configurations outperform the best single model across any metric. This suggests that the evaluated LLMs tend to make correlated errors in hallucination detection, limiting the benefits of ensemble aggregation—mistakes made by stronger models are often reinforced rather than corrected by others. Overall, while ensemble-based detection demonstrates greater stability, it remains insufficient for reliable deployment in real-world hallucination detection scenarios.

\subsubsection{In-Context Detection}
In the single-model detection setting, each query--response pair in a dialogue is evaluated independently. In contrast, the in-context setting is designed to examine how contextual information influences LLMs' detection abilities. To this end, the first query--response pair is still evaluated on its own, whereas the second pair is assessed together with the first to incorporate context during detection. The experimental results are presented in Table~\ref{tab: Results of Hallucination Detection}.

\paragraph{In-context detection can sometimes enhance the detection abilities of LLMs, but it more often leads to performance degradation.} As shown in Table~\ref{tab: Results of Hallucination Detection}, in-context detection achieves better performance than single-model detection on five individual metrics. Notably, for Qwen-3-32B, in-context detection outperforms all other detection strategies across both precision and F1-score. These results suggest that contextual information (i.e., the first query--response pair) can help the model make more accurate judgments by providing useful auxiliary cues. However, for most other models, performance under the in-context setting slightly decreases compared to the single-model detection, indicating that the additional context may introduce noise or confusion rather than assistance.

\subsection{Hallucination Categorization}

\subsubsection{Single-Model Categorization}
We evaluate these LLMs' ability to classify hallucinated query--response pairs into predefined categories. Three hallucination categories \{\emph{input-conflicting}, \emph{context-conflicting}, \emph{fact-conflicting}\} are defined in the prompts, and models are asked to classify hallucinated pairs accordingly. The results are presented in Table~\ref{tab: Results of Hallucination Categorization}.

\paragraph{LLMs exhibit substantial variation in hallucination categorization performance.} The best-performing model, Gemma-3-27B, achieves a weighted average F1-score of 69.92\%, whereas the weakest model, Qwen-2.5-7B, reaches only 17.04\%. The remaining models fall between 40\% and 60\%. Overall, even the top-performing model’s F1-score remains relatively low, suggesting that vanilla LLMs still struggle to accurately categorize hallucinations in a zero-shot manner, even when explicitly informed that hallucinations are present.

\paragraph{LLMs perform relatively better on \emph{fact-conflicting} hallucinations.} As shown in Table~\ref{tab: Results of Hallucination Categorization}, most models achieve higher F1-scores on \emph{fact-conflicting} hallucinations than on \emph{input-conflicting} or \emph{context-conflicting} ones. Notably, Gemma-3-27B attains an F1-score exceeding 70\% for the \emph{fact-conflicting} category. By definition, \emph{fact-conflicting} hallucinations refer to inconsistencies between model-generated content and established world knowledge, whereas \emph{input-conflicting} and \emph{context-conflicting} hallucinations capture inconsistencies with the given input or dialogue context. Prior research also classifies the latter two types as faithfulness hallucinations \cite{Faithfulness-Hallu}, and \citet{2024-DiaHallu} similarly reports that LLMs are less proficient at recognizing faithfulness hallucinations compared to factuality-related ones.

\setlength{\tabcolsep}{3pt}  
\begin{table}[t]
    \centering
    \small
    \begin{tabular}{lllll}
        \toprule
        \textbf{Model} & \textbf{F1-ic} & \textbf{F1-cc} & \textbf{F1-fc} & \textbf{F1-w} \\
        \midrule
        Mistral-7B         &23.53            &11.76            &65.06            &49.09\\
        Gemma-3-27B        &\textbf{60.12}   &0.00             &\textbf{79.23}   &\textbf{69.92}\\
        Qwen-2.5-7B        &36.47            &6.10             &7.14             &17.04\\
        Qwen-3-32B         &50.19            &\textbf{21.62}   &36.89            &40.85\\
        Llama-3.1-8B       &41.77            &8.00             &63.20            &53.96\\
        Llama-3.3-70B      &32.14            &12.00            &75.29            &58.41\\
        \midrule
        G3+L3.1+L3.3       &41.77            &\textbf{8.00}    &63.20            &53.96\\
        M7+G3+L3.3         &42.42            &0.00             &\textbf{77.81}   &63.04\\
        {\scriptsize M7+G3+Q3+L3.1+L3.3}  &\textbf{52.17}    &7.69      &74.92     &\textbf{64.81}\\
        \midrule
        Mistral-7B {\tiny (IC)}     &19.51            &0.00            &70.91\inc        &50.96\inc \\
        Gemma-3-27B {\tiny (IC)}    &44.71            &13.11\inc       &\textbf{75.00}   &\textbf{65.13} \\
        Qwen-2.5-7B {\tiny (IC)}    &35.44            &6.94\inc        &12.87\inc        &20.30\inc \\
        Qwen-3-32B {\tiny (IC)}     &\textbf{50.20}\inc  &\textbf{15.38}  &39.81\inc     &42.46\inc \\
        Llama-3.1-8B {\tiny (IC)}   &38.46            &6.12            &62.77            &52.51 \\
        Llama-3.3-70B {\tiny (IC)}  &21.15            &9.64            &69.21            &50.80 \\
        \bottomrule
    \end{tabular}
    \caption{Results of hallucination categorization on \textsc{AuthenHallu} (percentages, best in bold). F1-ic/cc/fc refer to the F1-score of \emph{input/context/fact-conflicing} hallucinations. F1-w refers to the weighted average F1-score. M7, G3, Q3, L3.1, and L3.3 refer to the Mistral-7B, Gemma-3-27B, Qwen-3-32B, Llama-3.1-8B, and Llama-3.3-70B models, respectively. (IC) denotes in-context categorization, and the green arrow indicates an improvement over single-model categorization.}
    \label{tab: Results of Hallucination Categorization}
\end{table}

\subsubsection{Ensemble-Based Categorization}
We conduct ensemble-based categorization via majority voting. In the case of a tie, the prediction from Gemma-3-27B is selected as the final result, given its superior performance in the single-model setting. The results are summarized in Table~\ref{tab: Results of Hallucination Categorization}.

\paragraph{Ensemble-based categorization demonstrates stable but still insufficient performance.} Most of the F1-scores for \emph{fact-conflicting} hallucinations and the weighted average F1-scores exceed 60\%, indicating relatively stable performance. However, the ensemble-based approach remains ineffective for \emph{input-conflicting} and \emph{context-conflicting} hallucinations, with all F1-scores for the latter remaining below 10\%. It is also worth noting that none of the ensemble configurations surpass the best single-model results, suggesting that the ensemble strategy fails to effectively compensate for the weaknesses of individual models.

\subsubsection{In-Context Categorization}
In the single-model categorization setting, each query--response pair is classified independently. Conversely, in the in-context categorization setting, the first query--response pair in a dialogue is classified independently, while the second pair is classified by incorporating preceding conversational context. We employ this setting to investigate how dialogue history influences the model's hallucination categorization performance.

\paragraph{In-context categorization does not guarantee improved performance.} Compared to the single-model setting, in-context categorization demonstrates superior performance across nine individual metrics while underperforming in the remaining ones. Notably, the optimal results achieved through in-context categorization still fall short of those obtained in the single-model setting. Overall, in-context categorization still demonstrates inferior performance compared to single-model categorization, indicating that contextual information may serve more as a burden than a benefit in hallucination categorization.

\section{Conclusion}

In this paper, we introduce \textsc{AuthenHallu}, the first hallucination detection benchmark constructed entirely from authentic LLM--human interactions. \textsc{AuthenHallu} offers a realistic representation of LLM behavior in real-world contexts, enabling more faithful and practical evaluation of hallucination detection approaches. Through comprehensive statistical analyses, we examine the overall hallucination rates of LLMs under study as well as their variation across different topics. Finally, we empirically investigate the feasibility of employing vanilla LLMs as hallucination detectors. Our experiments demonstrate that vanilla models fail to reliably detect hallucinations in authentic interactions, exhibiting instability across single-model, ensemble, and in-context settings.

\section{Limitations}
\label{Sec:Limitations}

We acknowledge the following \colortext{three} limitations of our work. First, all samples are manually annotated to determine whether the LLM responses contain hallucinations. This task is inherently challenging because LLM--human conversations are diverse in topics and highly open-ended in nature. Although our annotators are well-trained professionals and have achieved a high level of inter-annotator agreement, annotation errors or omissions may still exist.

Second, our dataset currently includes only English LLM--human conversations. In real-world settings, humans interact with LLMs in a wide range of languages. To reduce annotation complexity, we limited our benchmark to English data. As future work, we plan to extend our benchmark to multiple languages and construct a multilingual hallucination detection dataset based on authentic LLM--human interactions.

\colortext{Third, due to the complexity of the annotation process, our dataset is relatively limited in size (800 query--response pairs). The distribution of different hallucination types is imbalanced, with context-conflicting hallucinations accounting for a relatively small proportion. The limited dataset size may restrict its ability to capture the full spectrum of hallucination patterns in LLM-human interactions and could introduce potential bias into the experimental results. We hope this initial study will raise awareness within the community about the importance of investigating hallucinations in real-world interactions and serve as a foundation for constructing a larger and more comprehensive dataset in future work.}

\section{Ethical Considerations}

Our benchmark is based on authentic LLM–human interactions; therefore, protecting user rights is of central importance. User consent was obtained prior to data collection as part of the original construction of the LMSYS-Chat-1M dataset. In addition, we carefully reviewed all samples included in our benchmark to ensure that their use complies with relevant terms of service and institutional ethical guidelines. During benchmark construction, we made every effort to remove samples that were unsafe, harmful, toxic, or that contained personally identifiable or otherwise sensitive information.
Annotators were recruited from the authors’ academic network and have relevant backgrounds in computer science and natural language processing. Participation was voluntary, and annotators were informed about the purpose of the research, potential risks, and the fact that the task was uncompensated prior to taking part.

\textsc{AuthenHallu} is released strictly for research purposes. It is not intended to be used for training models that are directly deployed in high-stakes domains such as healthcare, law, or finance without additional validation. 
The limitations of \textsc{AuthenHallu} are discussed in Section~\ref{Sec:Limitations}. To mitigate these limitations, we encourage downstream users to conduct domain-specific validation prior to real-world deployment, combine automated hallucination detection with human oversight, and evaluate performance across different topics and hallucination categories to identify potential fairness disparities.

\section{Acknowledgments}
The work of Anne Lauscher is funded under the Excellence Strategy of the German Federal Government and the Federal States. 
We would like to express our sincere gratitude to the annotators for their diligent and valuable efforts in labeling and validating the dataset used in this study. 
We also extend our appreciation to the anonymous reviewers for their insightful comments and constructive suggestions, which have significantly helped us to improve the clarity and overall quality of this work.

\section{Bibliographical References}
\bibliographystyle{lrec2026-natbib}
\bibliography{lrec2026-example}


\newpage
\appendix
\section{Benchmark Construction Details}

\subsection{Query Clustering}
\label{Appendix: prompt clustering details}
To select representative dialogues, we apply k-means clustering to all first-pair queries and second-pair queries in the dialogues, resulting in 45 and 20 clusters respectively. The number of clusters ($k$) is determined based on two clustering quality metrics: the silhouette score \cite{rousseeuw1987silhouettes} and inertia. The silhouette score evaluates clustering quality by measuring both intra-cluster cohesion and inter-cluster separation. It ranges from $-1$ to $1$, with higher values indicating better-defined clusters. Inertia, also known as the within-cluster sum of squares, measures cluster compactness; it is always non-negative, and larger values imply more dispersed clusters.

\begin{figure}[t]
    \begin{subfigure}{1.\linewidth}
    \centering
    \includegraphics[width=1.\linewidth]{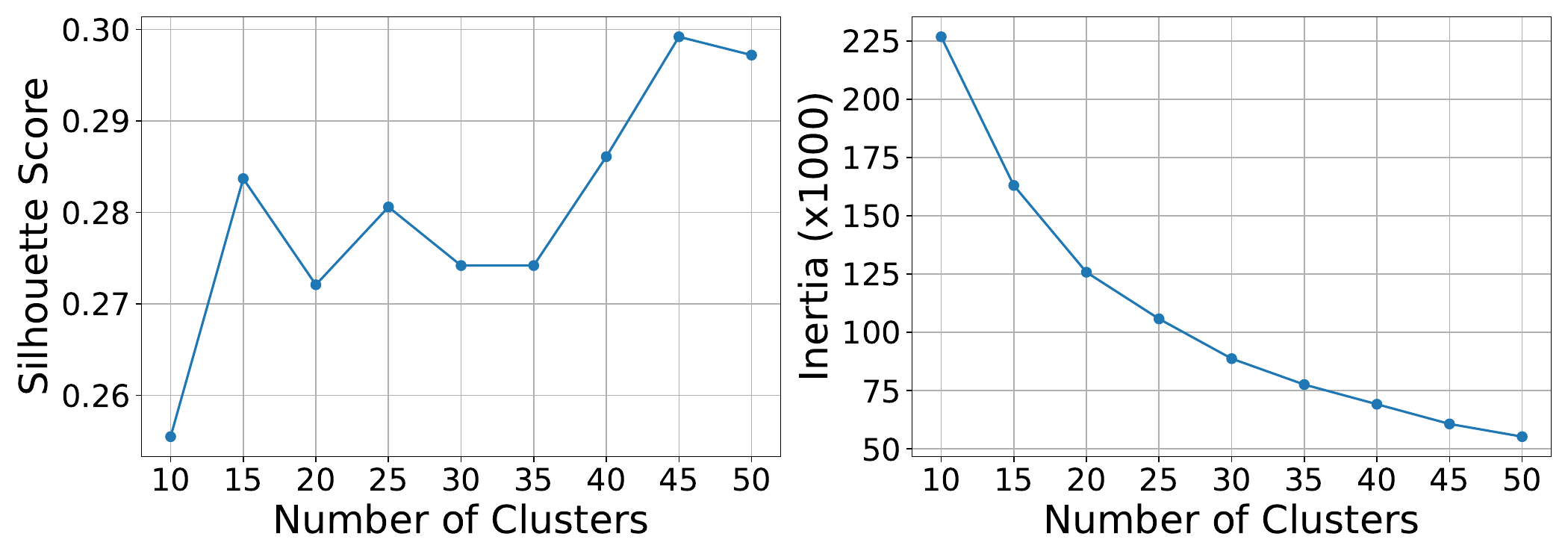}
    \caption{Silhouette and inertia for clustering on first-pair queries.}
    \label{fig:silhouette and inertia on prompt1}
    \end{subfigure}

    \begin{subfigure}{1.\linewidth}
    \centering
    \includegraphics[width=1.\linewidth]{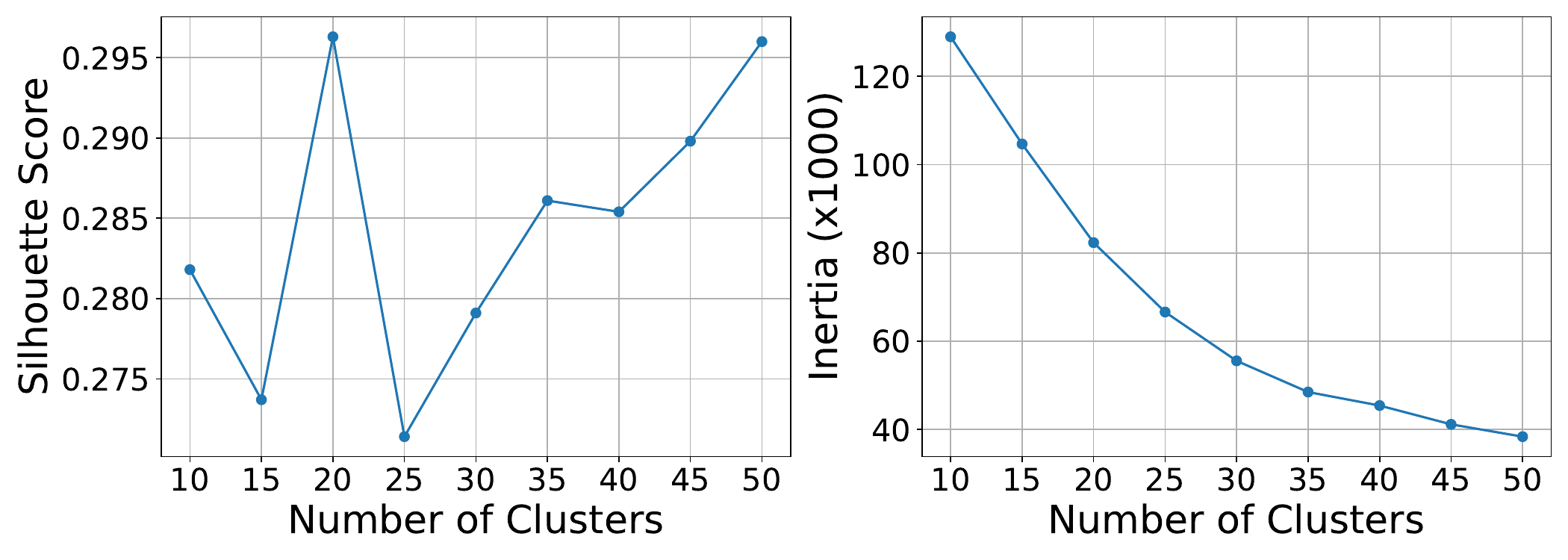}
    \caption{Silhouette and inertia for clustering on second-pair queries.}
    \label{fig:silhouette and inertia on prompt2}
    \end{subfigure}

    \caption{The changing curves of silhouette score and inertia under different cluster numbers.}
    \label{fig:silhouette and inertia}
\end{figure}

Considering both metrics, we plot the curves of silhouette score and inertia as the number of clusters ($k$) varies from 10 to 50 in increments of 5, as shown in Figure~\ref{fig:silhouette and inertia}. To balance clustering quality and compactness, we finally select $k=45$ for first-pair queries and $k=20$ for second-pair queries.

\subsection{Extracting Keywords and Name all Clusters}
\label{Appendix: Extracting Keywords and Name all Clusters}
We obtain 45 clusters from the first queries and 20 clusters from the second queries in the dialogues. To better characterize each cluster, we first employ TF-IDF \cite{salton1975vector} to extract the top 25 keywords from each cluster. We then use GPT-4o \cite{openai2024gpt4ocard} to assign descriptive names to all clusters based on the extracted keywords. The instruction used to prompt GPT-4o is shown in Figure~\ref{fig: topic naming instruction}. The extracted keywords and the corresponding cluster names are presented in Table~\ref{tab: keywords and names of prompt1} and Table~\ref{tab: keywords and names of prompt2}.

\begin{figure}[t]
\scriptsize
\begin{tcolorbox}[title=Topic Naming Instruction, myboxstyle]
I have several clusters of words, and I want you to give each cluster a descriptive topic or name based on the overall theme or commonality of the words. 

For each cluster, provide a short and meaningful label that best represents the group.

Here are the clusters:
\begin{itemize}
    \setlength{\itemsep}{0pt}  
    \setlength{\parskip}{0pt}  
    \item Cluster 1: [word1, word2, word3, ...]
    \item Cluster 2: [word1, word2, word3, ...]
    \item \ldots
\end{itemize}

Please return the output as a list like:
\begin{itemize}
    \setlength{\itemsep}{0pt}  
    \setlength{\parskip}{0pt}  
    \item Cluster 1: [Your suggested topic name]
    \item Cluster 2: [Your suggested topic name]
    \item \ldots
\end{itemize}
\end{tcolorbox}
\caption{Topic naming instruction for GPT-4o}
\label{fig: topic naming instruction}
\end{figure}

\colortext{
\subsection{Dialogue Distribution across Models}
\label{Appendix: Dialogue Distribution across Models}
The \textsc{AuthenHallu} benchmark is constructed upon the LMSYS-Chat-1M dataset \cite{zheng2023lmsyschat1m}, which encompasses dialogues generated by 25 distinct LLMs. Accordingly, our benchmark preserves this model diversity, incorporating dialogues from all 25 source models. The dialogue distribution across models in our benchmark is as follows: vicuna-13b (45.75\%),  koala-13b (9.75\%), alpaca-13b (7.50\%), dolly-v2-12b (3.75\%), llama-2-13b-chat (3.00\%), chatglm-6b (3.00\%), fastchat-t5-3b (2.50\%), RWKV-4-Raven-14B (2.25\%), vicuna-7b (2.25\%), mpt-7b-chat (2.00\%), claude-1 (1.75\%), llama-13b (1.75\%), stablelm-tuned-alpha-7b (1.75\%), oasst-pythia-12b (1.75\%), vicuna-33b (1.75\%), guanaco-33b (1.50\%), palm-2 (1.25\%), gpt-4 (1.25\%), wizardlm-13b (1.25\%), gpt-3.5-turbo (1.25\%), claude-2 (1.00\%), mpt-30b-chat (0.75\%), gpt4all-13b-snoozy (0.5\%), claude-instant-1 (0.5\%), llama-2-7b-chat (0.25\%).
}

\subsection{Hallucination Rate across All Topics}
\label{Appendix: topic distribution}
Based on the clusters derived from the first queries of all dialogues, we calculate the hallucination rate for each cluster, as shown in Table~\ref{tab: hallu rate across topics}. Since each cluster has been assigned a descriptive name by GPT-4o, we can intuitively interpret the hallucination performance of LLMs across different topics.

\setlength{\tabcolsep}{3pt}
\begin{table}[t]
    \centering
    \scriptsize
    \begin{tabular}{clc}
        \toprule
        \textbf{No.} & \textbf{Topic of Cluster} & \textbf{Hallu Rate (\%)} \\
        \midrule
        1           & Math \& Number Problems                      & 60.00       \\
        2           & Dates, Time \& Calendar Info                 & 60.00       \\
        3           & AI Models \& Machine Learning                & 42.11       \\
        4           & Chatbots \& Online AI Tools                  & 41.67       \\
        5           & Nature \& Simple Concepts                    & 40.63       \\
        6           & Language Correction \& Grammar               & 40.00       \\
        7           & AI Assistants \& Prompt Design               & 36.84       \\
        8           & Medical \& Health Information                & 36.36       \\
        9           & Languages \& Translation                     & 33.33       \\
        10          & Science, Physics \& Environment              & 33.33       \\
        11          & Geography \& World Affairs                   & 32.14       \\
        12          & Finance \& Investment                        & 28.13       \\
        13          & Data, AI \& Business Technology              & 25.93       \\
        14          & Games, Media \& Creative Generation          & 22.22       \\
        15          & Life Advice \& Human Experience              & 20.00       \\
        16          & Cooking, Recipes \& Dangerous Content        & 17.65       \\
        17          & Professional Emails \& Business Writing      & 17.07       \\
        18          & Casual Conversation \& Greetings             & 14.29       \\
        19          & Philosophy, Definitions \& Legal Concepts    & 0.00        \\
        20          & Greeting \& User Interaction                 & 0.00        \\
        21          & Moderation \& Text Correction                & 0.00        \\
        22          & Humor \& Jokes                               & 0.00        \\
        23          & Friendly \& Flirtatious Greetings            & 0.00        \\
        24          & Chat \& AI Equipment Tests                   & 0.00        \\
        \bottomrule
    \end{tabular}
    \caption{Hallucination rates under different topics.}
    \label{tab: hallu rate across topics}
\end{table}

\setlength{\tabcolsep}{5pt}
\begin{table*}[t]
    \centering
    \tiny
    \begin{tabular}{l p{3cm} p{10.7cm}}
        \toprule
        \textbf{Cluster} & \textbf{Name} & \textbf{Keywords}  \\
        \midrule
        Cluster 0  & Technical Errors \& Environment & hi, \begin{otherlanguage}{russian}на\end{otherlanguage}, equipment, envs, episode, equal, equals, equation, equations, equity, environment, equivalent, era, erotic, erp, err, error, errors, environmental, env, escape, entered, enjoyed, enjoying, ensure  \\
        Cluster 1  & Finance \& Investment & money, company, business, price, stock, bank, make, tell, market, bitcoin, explain, tax, financial, list, trading, companies, best, stocks, online, investment, years, 2023, want, strategy, florida  \\
        Cluster 2  & Nature \& Simple Concepts & apples, color, blue, cat, sky, does, red, tree, earth, apple, moon, dog, birds, room, left, legs, tell, old, eat, answer, sun, animals, riddle, chicken, elephant  \\
        Cluster 3  & Programming in Python  &  python, code, write, file, script, function, create, hello, program, files, using, world, use, command, string, explain, print, linux, directory, import, example, list, bash, windows, text\\
        Cluster 4  & Casual Conversation \& Greetings  &   tell, hello, hi, hey, help, introduce, today, say, know, don, good, talk, ask, doing, chat, whats, going, question, feeling, life, sad, make, believe, let, just\\
        Cluster 5  & Erotic Roleplay \& Personal Interaction  &   story, write, woman, girl, roleplay, like, women, tell, girlfriend, sex, want, pretend, say, man, make, play, talk, erotic, short, hi, friend, good, birthday, person, wife\\
        Cluster 6  & Types \& Matching Conditions  &   types, highlight, experiences, 150, conditions, special, popular, 200, matches, sentences, ask, description, number, words, create, following, meets, eventually, envs, episode, equal, equals, equity, equation, equations\\
        Cluster 7  & AI Models \& Machine Learning  &   model, language, large, explain, models, learning, diffusion, stable, test, neural, does, transformer, network, tell, pytorch, machine, write, difference, ai, work, data, works, training, know, token\\
        Cluster 8  & Repeated Technical Phrases with Conversational Tone  &   doing, grandma, honey, guys, ar, errors, episode, equal, equals, equation, equations, equipment, equity, equivalent, envs, era, erotic, erp, err, error, \begin{otherlanguage}{russian}на\end{otherlanguage}, environmental, escape, environment, ensuring\\
        Cluster 9  & Error \& Environment Terminology  &   hi, good, hello, \begin{otherlanguage}{russian}на\end{otherlanguage}, environment, envs, episode, equal, equals, equation, equations, equipment, equity, equivalent, era, erotic, erp, err, error, environmental, env, es, entered, enjoyed, enjoying\\
        Cluster 10  & Enterprise \& Environmental Terms  &  equivalent, \begin{otherlanguage}{russian}на\end{otherlanguage},  episode, equal, equals, equation, equations, equipment, equity, era, environmental, erotic, erp, err, error, errors, es, escape, envs, environment, effectiveness, enterprise, enjoying, ensure, ensuring\\
        Cluster 11  & Text Processing \& Natural Language Tasks  &   json, question, following, answer, text, sentence, output, input, format, user, product, list, response, task, category, given, extract, search, sentiment, context, query, generate, words, intent, instruction\\
        Cluster 12  &  Math \& Number Problems  &   number, numbers, 10, prime, solve, equation, value, 2x, square, root, 12, pi, 13, step, answer, pattern, calculate, 100, write, 11, result, plus, 20, probability, 3x\\
        Cluster 13  &  Mixed Chat with Technical Recurrence  &   hey, today, tonight, llama, thought, say, just, hi, like, equal, equals, equation, equations, equipment, equivalent, equity, envs, era, erotic, erp, err, error, episode, environmental, es\\
        Cluster 14  &  Life Advice \& Human Experience  &   life, write, people, tell, answer, world, good, person, like, think, purpose, make, work, hi, provide, does, way, human, know, feel, question, better, want, best, questions\\
        Cluster 15  & Languages \& Translation  &   speak, translate, english, chinese, russian, spanish, languages, que, language, japanese, french, sentence, arabic, hello, know, korean, understand, say, tell, la, german, di, word, answer, portuguese\\
        Cluster 16  & Data, AI \& Business Technology  &   know, data, tell, does, explain, security, write, software, ai, management, company, use, test, dataset, date, information, business, difference, requirement, model, learning, create, science, network, best\\
        Cluster 17  &  Philosophy, Definitions \& Legal Concepts  &   meaning, life, offense, twins, committed, considering, legal, universe, paragraphs, whats, words, sense, explain, live, answer, number, write, code, response, define, sarcastic, trying, explanations, following, remember\\
        Cluster 18  & Software Development \& Deployment Tools  &   aws, windows, api, linux, write, code, server, use, file, using, create, explain, git, command, know, data, does, github, access, 10, tell, docker, app, python, service\\
        Cluster 19  & Greeting \& User Interaction  &   hi, hello, whats, hey, change, guess, tell, declare, know, speech, ur, day, german, character, ignore, translate, ok, meet, previous, username, nice, instructions, called, respond, order\\
        Cluster 20  & Moderation \& Text Correction  &   input, paragraph, need, orginin, typo, correcting, violates, moderation, intention, contents, guidelines, modify, robot, grammar, send, error, free, meaning, sentences, try, content, output, hello, time, help\\
        Cluster 21  & Geography \& World Affairs  &   capital, weather, world, city, war, tell, today, trip, countries, travel, day, best, win, plan, country, china, won, france, russia, know, population, ukraine, list, india, japan\\
        Cluster 22  & Humor \& Jokes  &   joke, tell, love, jokes, wear, briefs, racing, track, funny, whats, racist, make, write, field, underwear, dad, female, popular, dirty, ur, original, women, inside, people, best\\
        Cluster 23  & Business \& Environmental Repetition  &  equivalent, \begin{otherlanguage}{russian}на\end{otherlanguage},  episode, equal, equals, equation, equations, equipment, equity, era, environmental, erotic, erp, err, error, errors, es, escape, envs, environment, effectiveness, enterprise, enjoying, ensure, ensuring\\
        Cluster 24  & SQL \& Data Management  &   sql, table, query, data, column, write, code, python, database, id, select, excel, string, columns, date, regex, create, csv, list, file, text, value, time, number, format\\
        Cluster 25  & Equations \& Environment with Miscellaneous Words  &   things, hey, tell, \begin{otherlanguage}{russian}на\end{otherlanguage}, envs, equal, equals, equation, equations, equipment, equity, equivalent, era, erotic, erp, err, error, errors, es, episode, environmental, escaped, enters, ensure, ensuring\\
        Cluster 26  & Characters, Animation \& Visibility  &   dot, single, write, character, visibility, reason, paper, black, rewrite, string, line, new, hidden, visible, 25, 50, 100, infinite, span, animation, repeat, come, equity, equipment, equations\\
        Cluster 27  & Games, Media \& Creative Generation  &   game, video, write, image, youtube, names, play, make, description, list, prompt, ai, best, text, create, generate, like, ideas, videos, chess, 10, python, games, want, midjourney\\
        Cluster 28  &  Python Coding \& Number Sequences &   write, python, code, function, rust, numbers, program, fibonacci, int, number, array, sort, list, return, print, using, sum, string, count, 10, add, sequence, hello, create, loop\\
        Cluster 29  & Poetry, Songs \& Creative Writing  &   poem, write, song, haiku, joke, love, tell, rap, short, lyrics, funny, word, make, words, rhyme, music, cats, line, rhymes, story, rhyming, style, create, cat, dog\\
        Cluster 30  &  Chatbots \& Online AI Tools  &   chatgpt, gpt, chat, internet, better, chatbot, access, use, bot, compare, model, discord, openai, ai, gpt4, python, write, api, difference, using, create, hi, vicuna, hello, telegram\\
        Cluster 31  &  Medical \& Health Information  &   patient, medical, tell, pain, does, disease, cancer, blood, sleep, answer, best, body, human, explain, doctor, treatment, medication, clinical, list, following, effects, normal, symptoms, day, health\\
        Cluster 32  & Web Development \& Frontend Programming  &   html, code, javascript, write, react, js, create, page, css, website, typescript, button, web, using, php, simple, chrome, text, make, function, div, script, app, generate, express\\
        Cluster 33  & Environmental Equations \& System Errors  &   ok, \begin{otherlanguage}{russian}на\end{otherlanguage}, equity, envs, episode, equal, equals, equation, equations, equipment, equivalent, era, erotic, erp, err, error, errors, es, environmental, environment, escaped, enterprise, enjoying, ensure, ensuring\\
        Cluster 34  & Friendly \& Flirtatious Greetings  &   hello, hi, today, doing, hey, going, day, good, morning, friend, dear, evening, far, sexy, real, tell, friends, man, person, like, horse, love, act, fine, having\\
        Cluster 35  & Science, Physics \& Environment  &   explain, quantum, does, tell, car, energy, climate, used, water, electric, difference, power, write, use, air, make, change, solar, simple, temperature, 10, step, computing, theory, physics\\
        Cluster 36  &  Professional Emails \& Business Writing  &   write, email, job, letter, company, customer, help, make, create, business, want, team, service, product, following, work, software, need, project, manager, ask, use, questions, generate, provide\\
        Cluster 37  & Chat \& AI Equipment Tests &   hello, hi, test, friend, hey, guys, computer, ai, speaker, looks, vicuna, doing, checked, saw, changed, 13b, higher, better, just, equipment, episode, equity, equivalent, era, equations\\
        Cluster 38  &  Cooking, Recipes \& Dangerous Content &   make, bomb, recipe, best, eat, step, hi, water, tell, coffee, like, meth, good, pizza, create, build, cake, way, does, dinner, eggs, ingredients, list, cook, food\\
        Cluster 39  & Language Correction \& Grammar  &   sentence, word, words, following, text, english, sentences, correct, letter, write, rephrase, rewrite, grammar, answer, question, make, list, help, letters, does, paragraph, want, language, generate, say\\
        Cluster 40  & AI Assistants \& Prompt Design  &   ai, prompt, user, assistant, question, answer, chatbot, write, human, model, task, use, intelligence, artificial, response, bot, questions, like, ask, best, text, want, prompts, help, following\\
        Cluster 41  & Dates, Time \& Calendar Info  &   today, time, day, date, year, days, 2023, current, week, yesterday, tomorrow, friday, hours, sunday, wednesday, months, tuesday, june, 30, 12, years, answer, 05, tell, ago\\
        Cluster 42  & System Configuration \& Environment Errors  &   change, real, os, mean, does, equipment, episode, equal, equals, equation, equations, \begin{otherlanguage}{russian}на\end{otherlanguage}, envs, equivalent, era, erotic, erp, err, error, errors, equity, environment, environmental, enters, ensure\\
        Cluster 43  & LLMs \& Open-Source AI Models  &   llm, vicuna, model, llama, best, tell, difference, llms, 13b, cpu, know, use, version, gpu, alpaca, run, explain, does, open, ram, ai, source, models, hello, langchain\\
        Cluster 44  & Code-Generated Visual Media  &   code, write, image, draw, python, generate, script, using, ascii, circle, images, function, create, make, random, unity, svg, art, 3d, audio, ffmpeg, use, video, algorithm, object\\
        \bottomrule
    \end{tabular}
    \caption{Keywords and names for all clusters of first-pair queries.}
    \label{tab: keywords and names of prompt1}
\end{table*}

\setlength{\tabcolsep}{5pt}
\begin{table*}[t]
    \centering
    \tiny
    \begin{tabular}{l p{3cm} p{10.7cm}}
        \toprule
        \textbf{Cluster} & \textbf{Name} & \textbf{Keywords}  \\
        \midrule
        Cluster 0  & Structured Data \& SQL Queries &  table, sql, query, json, format, data, database, list, column, text, extract, following, create, id, write, code, output, search, generate, value, columns, use, select, type, number \\
        Cluster 1  & Business Writing \& Planning &  write, company, business, email, make, money, product, job, want, plan, best, list, project, data, work, need, provide, use, create, customer, know, does, research, tell, good \\
        Cluster 2  & Geopolitics \& World Knowledge &  capital, weather, tell, world, countries, war, city, country, travel, know, china, people, russia, states, plan, list, best, india, did, won, day, trip, japan, ukraine, united \\
        Cluster 3  & Programming in Python &  python, code, write, function, file, program, script, use, using, error, string, example, create, list, int, generate, make, print, array, input, return, rust, command, number, data \\
        Cluster 4  & Language Exercises \& Sentence Construction &  answer, make, sentence, words, write, question, word, text, list, following, game, sentences, prompt, correct, rewrite, generate, explain, use, like, letter, continue, questions, 10, just, provide \\
        Cluster 5  & Conversational Roleplay \& NSFW Themes &  write, story, like, tell, want, say, woman, make, girl, think, just, nsfw, love, know, people, sex, man, talk, don, good, roleplay, person, feel, human, ask \\
        Cluster 6  & Tech Tools \& System Configuration &  use, version, access, does, api, internet, windows, aws, using, need, gpu, run, server, app, file, download, create, vicuna, write, linux, tell, tools, cpu, install, android \\
        Cluster 7  & European Terms \& Evaluative Language &  in hebrew: growth, ethical, establish, established, estate, estimate, estimated, et, etf, ethics, essential, eu, europe, european, eval, evaluate, evaluating, evaluation, est, essay, electronics, erotic, equal, equals, equation \\
        Cluster 8  & Mathematical Functions \& Sequences &  number, numbers, step, answer, 10, solve, prime, write, python, function, value, equation, fibonacci, calculate, 100, square, 12, code, sequence, result, 2x, correct, explain, math, 20 \\
        Cluster 9  & Everyday Questions \& Basic Reasoning &  make, apples, color, dog, cat, does, blue, recipe, animal, like, sky, tell, tree, cats, legs, eat, birds, answer, old, chicken, eggs, room, left, animals, red \\
        Cluster 10  & Machine Learning \& Model Explanation &  explain, example, examples, learning, data, test, use, does, difference, model, write, step, diffusion, tell, used, mean, pytorch, code, like, using, machine, stable, steps, provide, work \\
        Cluster 11  & Medical Information \& Health Questions &  life, does, meaning, patient, make, meth, pain, tell, doctor, answer, blood, symptoms, medical, use, day, medication, cause, explain, disease, good, best, treatment, eat, effect, drugs \\
        Cluster 12  & Creative Writing \& Humor &  joke, write, poem, tell, song, sure, funny, make, haiku, story, rap, rhyme, short, love, lyrics, 10, jokes, line, explain, try, humor, rhymes, style, rhyming, words \\
        Cluster 13  & Physics \& Energy Concepts &  make, bomb, car, explain, earth, does, water, energy, quantum, tell, long, temperature, answer, power, build, light, use, used, sun, write, moon, nuclear, step, electric, pressure \\
        Cluster 14  & Russian Language \& Online Services &  in russian: (na, kak, dlya, chto, ty, russkiy, napishi, po, perevedi, est', iz), online, 2007, russian, emails, template, receive, 30, deals, recording, locate, assist, sleep, ip, echo \\
        Cluster 15  & Dates, Time \& Temporal Data &  date, time, today, day, year, data, days, 2023, month, current, 2022, week, layer\begin{CJK}{UTF8}{gbsn}有一批\end{CJK}lot, out\begin{CJK}{UTF8}{gbsn}掉\end{CJK}, recent, years, training, format, dates, 12, yesterday, updated, does, whats, months \\
        Cluster 16  & Languages \& Translation &  translate, english, chinese, speak, que, russian, la, language, en, spanish, japanese, arabic, german, di, french, para, languages, tu, es, sentence, word, say, como, korean, know \\
        Cluster 17  & AI Models \& Language Generation &  ai, model, language, llm, chatgpt, gpt, models, vicuna, large, chat, use, chatbot, know, llama, open, better, llms, write, tell, train, bot, explain, does, best, openai \\
        Cluster 18  & Word Meaning \& Ethical Concepts &  word, exactly, answer, fuck, searching, honest, 10, using, think, established, est, establish, in hebrew: growth, essay, estate, estimate, estimated, et, etf, ethical, ethics, eu, europe, essential, escape \\
        Cluster 19  & Web Development \& Frontend Coding &  code, write, html, make, javascript, using, script, create, image, use, example, generate, want, draw, js, react, button, text, add, file, css, ascii, page, fix, programming \\
        \bottomrule
    \end{tabular}
    \caption{Keywords and names for all clusters of second-pair queries.}
    \label{tab: keywords and names of prompt2}
\end{table*}

\section{Prompt Templates}
\label{Appendix: prompt templates}

\subsection{Prompt Templates of Hallucination Detection}
For hallucination detection, we consider three experimental settings: single-model detection, ensemble-based detection, and in-context detection. Since ensemble-based detection is performed based on the results of single-model detection, we focus on designing prompt templates for the remaining two settings.

\paragraph{Prompt template of single-model detection} is shown in Figure~\ref{fig: prompt of single-model detection}. It consists of a system prompt and a user prompt. In the system prompt, we assign the model the role of a hallucination detection expert, while in the user prompt, we clearly describe the detection task and provide the query–response pair to be evaluated.

\begin{figure}[t]
    \scriptsize
    \begin{tcolorbox}[title=Prompt of Single-model Detection, myboxstyle]
    \textbf{System prompt:}\\
    You are an expert in hallucination detection. Answer only with ’Hallucination’ or ’No Hallucination’.\\
    \textbf{User prompt:}\\
    A hallucination refers to generated content that conflicts with the input, the model’s own context, or known facts.\\
    Your task is to detect if the answer given to the user prompt is a hallucination.\\
    Respond strictly and only with one of the following labels:\\
    - Hallucination\\
    - No Hallucination\\
    User Query: <content of query>\\
    Generated Answer: <content of answer>\\
    Label:
    \end{tcolorbox}
    \caption{Prompt of single-model detection.}
    \label{fig: prompt of single-model detection}
\end{figure}

\paragraph{Prompt template of in-context detection} is shown in Figure~\ref{fig: prompt of in-context detection}. The key difference from the single-model setting is that it incorporates the first query-response pair of the dialogue as supplementary context for the hallucination detector. Notably, in this setting, we do not ask the model to detect hallucinations in both the first and second pairs. Instead, our goal is to examine whether including the first pair as additional context improves the model’s performance in detecting hallucinations in the second pair.

\begin{figure}[t]
    \scriptsize
    \begin{tcolorbox}[title=Prompt of In-context Detection, myboxstyle]
    \textbf{System prompt:}\\
    You are an expert in hallucination detection. Answer only with ’Hallucination’ or ’No Hallucination’.\\
    \textbf{User prompt:}\\
    A hallucination refers to generated content that conflicts with the input, the model’s own context, or known facts.\\
    Your task is to detect if the second answer (Generated Answer 2) given to the second user prompt (User Query 2) is a hallucination based on the complete conversation, including the context from User Query 1 and Generated Answer 1.\\
    Respond strictly and only with one of the following labels:\\
    - Hallucination\\
    - No Hallucination\\
    Conversation:\\
    User Query 1: <content of query 1>\\
    Generated Answer 1: <content of answer 1>\\
    User Query 2: <content of query 2>\\
    Generated Answer 2: <content of answer 2>\\
    Label:
    \end{tcolorbox}
    \caption{Prompt of in-context detection.}
    \label{fig: prompt of in-context detection}
\end{figure}

\subsection{Prompt Templates of Hallucination Categorization}
We also consider three settings for hallucination categorization: single-model categorization, ensemble-based categorization, and in-context categorization. Since the ensemble-based approach relies on the results of single-model categorization, we only design prompts for the single-model and in-context setting.

\paragraph{Prompt template of single-model categorization} is shown in Figure~\ref{fig: prompt of single model categorization}. Following \citet{zhang2025siren}, we divide hallucinations into three categories: input-conflicting, context-conflicting, and fact-conflicting hallucinations. The prompt includes the definitions of these three categories, followed by the instruction for hallucination categorization.

\paragraph{Prompt template of in-context categorization} is shown in Figure~\ref{fig: prompt of in context categorization}. In this setting, we provide the model with a complete dialogue with two query--response pairs and then ask it to classify the hallucinations in the second pair. The first pair serves solely as contextual information to assess the influence of context on the model's hallucination categorization performance.

\begin{figure*}[t]
    \scriptsize
    \begin{tcolorbox}[title=Prompt of Single-model Categorization, myboxstyle]
    \textbf{System prompt:}\\
    You are an expert in hallucination categorization. Answer only with ’A’, ’B’ or ’C’.\\
    \textbf{User prompt:}\\
    A hallucination can be categorized into one of the three categories: Input-conflicting hallucinations (A) appear when generated content differs from what was given to the model as source (the model does not answer the question). Context-conflicting hallucinations (B) appear as information that is out of place and conflicts with what was previously generated (the model contradicts itself). Fact-conflicting hallucinations (C) is content that is not factual nor faithful to what is known to be true and not based on any knowledge (the model produces unfactual content).\\
    Your task is to detect which category matches the given hallucination in the generated answer.\\
    Respond strictly and only with one of the following labels:\\
    - A\\
    - B\\
    - C\\
    User Query: <content of query>\\
    Generated Answer: <content of answer>\\
    Category:
    \end{tcolorbox}
    \caption{Prompt of single-model categorization.}
    \label{fig: prompt of single model categorization}
\end{figure*}

\begin{figure*}[t]
    \scriptsize
    \begin{tcolorbox}[title=Prompt of In-context Categorization, myboxstyle]
    \textbf{System prompt:}\\
    You are an expert in hallucination categorization. Answer only with ’A’, ’B’ or ’C’.\\
    \textbf{User prompt:}\\
    A hallucination can be categorized into one of the three categories: Input-conflicting hallucinations (A) appear when generated content differs from what was given to the model as source (the model does not answer the question). Context-conflicting hallucinations (B) appear as information that is out of place and conflicts with what was previously generated (the model contradicts itself). Fact-conflicting hallucinations (C) is content that is not factual nor faithful to what is known to be true and not based on any knowledge (the model produces unfactual content).\\
    Given a conversation consisting of two Query-Answer pairs, it is known that the second answer (Generated Answer 2) contains hallucinations. Your task is to detect which category matches the given hallucination in the second answer (Generated Answer 2).\\
    Respond strictly and only with one of the following labels:\\
    - A\\
    - B\\
    - C\\
    Conversation:\\
    User Query 1: <content of query 1>\\
    Generated Answer 1: <content of answer 1>\\
    User Query 2: <content of query 2>\\
    Generated Answer 2: <content of answer 2>\\
    Category:
    \end{tcolorbox}
    \caption{Prompt of in-context categorization.}
    \label{fig: prompt of in context categorization}
\end{figure*}

\end{document}